# An Efficient Protocol for Negotiation over Combinatorial Domains with Incomplete Information

**Minyi Li**
Faculty of ICT
Swinburne University
Hawthorn, VIC 3122, Australia

**Quoc Bao Vo**
Faculty of ICT
Swinburne University
Hawthorn, VIC 3122, Australia

**Ryszard Kowalczyk**
Faculty of ICT
Swinburne University
Hawthorn, VIC 3122, Australia

## Abstract

We study the problem of agent-based negotiation in combinatorial domains. It is difficult to reach optimal agreements in bilateral or multi-lateral negotiations when the agents' preferences for the possible alternatives are not common knowledge. Self-interested agents often end up negotiating inefficient agreements in such situations. In this paper, we present a protocol for negotiation in combinatorial domains which can lead rational agents to reach optimal agreements under incomplete information setting. Our proposed protocol enables the negotiating agents to identify efficient solutions using distributed search that visits only a small subspace of the whole outcome space. Moreover, the proposed protocol is sufficiently general that it is applicable to most preference representation models in combinatorial domains. We also present results of experiments that demonstrate the feasibility and computational efficiency of our approach.

## 1 Introduction

Multi-issue negotiation is one of the most preferred approaches for resolving conflicts in agent society [4]. It is being increasingly used in different multi-agent domains including trading systems, resource allocation, service level agreement negotiations, etc. [4, 9]. When multiple issues are involved in negotiation simultaneously, the agents with divergent preferences can cooperate to reach agreements that are beneficial for each other. But when the preferences of the participating agents are not common knowledge, they often fail to explore win-win possibilities and end up with inefficient results. Therefore, there is a need for negotiation protocols which can lead rational agents to optimal agreements. By optimal or efficient agreement, we refer to an agreement which is Pareto-optimal (or Pareto-efficient) [7]. An alternative or outcome is Pareto-optimal (or Pareto-efficient) if there exists no other alternative which is at least as good as this alternative for all agents and strictly better for at least one agent.

In this paper, we focus on the negotiation problems where the space of alternatives has a combinatorial structure [1]. For example, negotiations over multiple indivisible goods or resources (where the number of bundles an agent may obtain is exponential in the number of goods or resources). When the negotiating agents know about the preferences of each other, they can reach an efficient agreement using distributed protocols like one-step monotonic concession protocol or monotonic concession protocol [8], where each agent searches the entire space of possible agreements. Similar scenarios of multi-attribute decision-making with complete information have also been studied in the field of collective decision-making in combinatorial domains, i.e. voting theory and preference aggregation (for instance, [5], [7], [10] etc..), which determines either one, some, or all optimal alternatives from a given collection of the agents preferences according to a given preference aggregation rule. However, most negotiations in real life take place under incomplete information where the agents do not have complete knowledge about the preferences of the opponents. Some protocols for negotiation over multiple indivisible resources in incomplete information scenarios have been proposed by Brams and Taylor [2]. These protocols can produce optimal agreements only for negotiating over multiple uncorrelated resources, i.e., situations where the utility of possessing two resources is the sum of the utilities of possessing each individual resource. The scenarios they considered are similar to the negotiation problem in combinatorial domains

---
[1] The number of possible alternatives is exponential in the number of attributes that are involved in negotiation

where the attributes are all independent. But real-life negotiations typically involve inter-dependent attributes and the decision-making process tends to become much more complex. As an example, a research group plan to order several PCs and the group members need to decide on a standard group PC configuration. The decisions are not independent, because, perhaps, the preferred operating systems may depend on the given processor type. For instance, "I prefer to choose WinXP operating system rather than Linux if an Intel processor is given." Hence, we cannot decide on the issues separately. However, in the situations of multiple interdependent or correlated issues, these existing protocols can produce very inefficient agreements in negotiation[9].

Our objective is to design an efficient protocol for agent-based negotiation in combinatorial domains, which can lead participating agents to *Pareto-optimal* agreements. We consider a completely uncertain negotiation scenario where participating agents do not have any knowledge about the preferences of the other agents; and the agents do not want to reveal their preferences for the possible alternatives during the process of negotiation. We propose a two phase negotiation protocol `POANCD` (Protocol to reach Optimal agreement in Negotiation over Combinatorial Domains). The first phase of `POANCD` involves an iterative negotiation process to generate a set of initial agreements that are close to optimal. And then the second phase further enhances the initial agreements to be Pareto-optimal by searching for possible mutually beneficial agreements.

Our proposed protocol makes a distinct contribution from other existing voting protocols or aggregation mechanisms in the sense that it is under purely incomplete information setting and distributed manner. Moreover, our protocol differs from most of the existing research in the field of utility-based negotiation. Because it can not only handle quantitative preferences, but also works with purely qualitative preference models. It is general enough to allow for a various types of preferences and representation models in combinatorial domains. The preferences can be cardinal (e.g. utilities) or ordinal preferences (preference relations). And the representation models can be based on conditional preferences, for instance CP-nets [1] and its variants, which consist of a structural part that expresses the links between variables, and a "table" part containing the local preferences; or it can be based on propositional logic (or possibly a fragment of it), for instance prioritized goals, distance-based goals, weighted goals, bidding languages for combinatorial auctions. Another advantage of our proposed protocol is that each agent is required to consider only a small subset of alternatives instead of the entire outcome space. It requires significantly less outcome comparisons compared to exhaustive search in most negotiation instances.

## 2 Preliminaries

### 2.1 Combinatorial domains

Let $\mathbf{V} = \{X_1, \ldots, X_m\}$ be a set of $m$ attributes in a combinatorial domain, For each $X \in \mathbf{V}$, $D(X)$ is the *domain* of $X$. A variable $X$ is *binary* if $D(X) = \{x, \bar{x}\}$. An alternative is uniquely identified by the combination of its attribute values. Hence, there are $D(X_1) \times \cdots \times D(X_m)$ possible alternatives (outcomes), denoted by $O$. Elements of $O$ are denoted by $o$, $o'$, $o''$ etc. and represented by concatenating the values of the variables. For example, if $\mathbf{V} = \{A, B, C\}$, $D(A) = \{a, \bar{a}\}$, $D(B) = \{b, \bar{b}\}$ and $D(C) = \{c, \bar{c}\}$, then the assignment $a\bar{b}c$ assigns $a$ to variable $A$, $\bar{b}$ to $B$ and $c$ to $C$. If $\mathbf{X} = \{X_{\sigma_1}, \ldots, X_{\sigma_\ell}\} \subseteq \mathbf{V}$, with $\sigma_1 < \cdots < \sigma_\ell$ then $D(\mathbf{X})$ denotes $D(X_{\sigma_1}) \times \cdots \times D(X_{\sigma_\ell})$ and $\mathbf{x}$ ($\mathbf{x} \in D(\mathbf{X})$) denotes an assignment of variable values to $\mathbf{X}$. If $\mathbf{X} = \mathbf{V}$, $\mathbf{x}$ is a *complete assignment* (corresponds to a possible outcome); otherwise $\mathbf{x}$ is called a *partial assignment*. If $\mathbf{x}$ and $\mathbf{y}$ are assignments to disjoint sets $\mathbf{X}$ and $\mathbf{Y}$, respectively ($\mathbf{X} \cap \mathbf{Y} = \emptyset$), we denote the combination of $\mathbf{x}$ and $\mathbf{y}$ by $\mathbf{xy}$. If $\mathbf{X} \cup \mathbf{Y} = \mathbf{V}$, we call $\mathbf{xy}$ a completion of assignment $\mathbf{x}$. We denote by $Comp(\mathbf{x})$ the set of completions of $\mathbf{x}$. For any assignment $\mathbf{x} \in D(\mathbf{X})$, we denote by $\mathbf{x}[X]$ the value $x \in D(X)$ ($X \in \mathbf{X}$) assigned to variable $X$ by that assignment; and $\mathbf{x}[\mathbf{W}]$ denotes the assignment of variable values $\mathbf{w} \in D(\mathbf{W})$ assigned to the set of variables $\mathbf{W} \subseteq \mathbf{X}$ by that assignment. We allow concatenations of partial assignments. For instance, let $\mathbf{V} = \{A, B, C, D, E, F\}$, $\mathbf{X} = \{A, B, C\}$, $\mathbf{Y} = \{D, E\}$, $\mathbf{x} = a\bar{b}\bar{c}$, $\mathbf{y} = \bar{d}e$, then $\mathbf{xy}\bar{f}$ denotes the alternative $a\bar{b}\bar{c}\bar{d}e\bar{f}$.

### 2.2 Preference in combinatorial domains

Preference indicates the ranking (or order, precedence) of possible alternatives based on the satisfaction they could provide for an agent. In decision-making theory, the standard way to model the decision maker's preference is with his preference relations, also called a binary relation [3]. Let $o$, $o'$ be two possible outcomes $o, o' \in O$. Then '$\succeq$' is a preference relation on $O$ such that $o \succeq o'$ if and only if $o$ is at least as preferable as $o'$ (or, $o$ is weakly preferred to $o'$). And $o$ is strictly preferred to $o'$ (notation $o \succ o'$) if and only if $o \succeq o'$ but $o' \not\succeq o$. When $o \succeq o'$ and $o' \succeq o$, we say that the agent is indifferent between these two outcomes, denoted by $o' \sim o$. Moreover, in combinatorial domains, two out-comes $o$ and $o'$ can also be incomparable for an agent when $o \not\succeq o'$ and $o' \not\succeq o$, denoted by $o \bowtie o'$.

Given a problem over a combinatorial domain, the direct assessment of the preference relations between alternatives is usually infeasible due to the exponential number of alternatives. AI researchers have been developing languages for representing preferences on such domains in a succinct way, exploiting structural properties such as conditional preferential independence.

**Preferential Independency**

Let $\mathbf{X}$, $\mathbf{Y}$, and $\mathbf{Z}$ be nonempty sets that partition $\mathbf{V}$ and $\succ$ a preference relation over $D(\mathbf{V})$. $\mathbf{X}$ is *(conditionally) preferentially independent* of $\mathbf{Y}$ given $\mathbf{Z}$ iff for all $\mathbf{x}, \mathbf{x}' \in D(\mathbf{X})$, $\mathbf{y}, \mathbf{y}' \in D(\mathbf{Y})$, $\mathbf{z} \in D(\mathbf{Z})$:

$$\mathbf{xyz} \succ \mathbf{x'yz} \text{ iff } \mathbf{xy'z} \succ \mathbf{x'y'z}$$

Among those preference representation languages in combinatorial domain, CP-net (Conditional Preference Network) is one of the most studied languages.

**CP-nets**

A CP-net $\mathcal{N}$ [1] over a set of domain attribute $\mathbf{V} = \{X_1, \ldots, X_m\}$ is an annotated directed graph $\mathcal{G}$, in which nodes stand for the problem attributes. Each node $X$ is annotated with a conditional preference table $CPT(X)$, which associates a total order $\succ^{X|\mathbf{u}}$ with each instantiation $\mathbf{u}$ of $X$'s parents $Pa(X)$, i.e. $\mathbf{u} \in D(Pa(X))$. For instance, let $\mathbf{V} = \{A, B, C\}$, all three being binary, and assume that the preference of a given agent over all possible outcomes can be defined by a CP-net whose structural part is the directed acyclic graph $\mathcal{G} = \{(C, A), (C, B), (A, B)\}$; this means that the agent's preference over the values of $C$ is unconditional, preference over the values of $A$ (resp. $B$) is fully determined given the value of $C$ (resp. the values of $C$ and $A$). The preference statements contained in the conditional preference tables are written with the usual notation, that is, $\bar{a}c : \bar{b} \succ b$ means that when $A = \bar{a}$ and $C = c$, then $B = \bar{b}$ is preferred to $B = b$ (see for example Figure 1(a)).

For clarity of presentation, we attempt to describe our negotiation protocol with acyclic CP-nets, i.e., the relation graph does not contain circle. *However, the proposed protocol can be used to handle various preferences models and languages given the corresponding techniques for answering dominance queries[2] and outcome optimization queries[3]*. Notice that the negotiation process is elicitation-free: the agents are never asked to report their preferences. Therefore, the agents are not going to reveal their CP-nets and the structures of their CP-nets do not play any role.

---
[2]A dominance query, given two alternatives $o$ and $o'$, asks whether $o$ is preferred to $o'$ with respect to an agent's preferences.

[3]An outcome optimization query determines the set

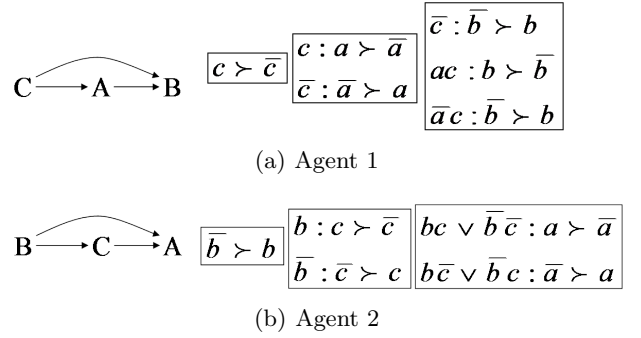

(a) Agent 1

(b) Agent 2

Figure 1: Two agents' CP-nets

## 3 Our proposed negotiation protocol

In this section we present our proposed protocol: <u>P</u>rotocol to reach <u>O</u>ptimal <u>a</u>greement in <u>N</u>egotiation over <u>C</u>ombinatorial <u>D</u>omains (POANCD). Before we go into technical detail, we first extend the concept of Pareto-optimality and rational agent in negotiation over combinatorial domains with incomparability.

**Definition 1** (Pareto-optimality). *An outcome $o$ is Weakly Pareto-optimal (WPO) if there exists no other outcome $o'$ such that all agents strictly prefers $o'$ to $o$. An outcome $o$ is Pareto-optimal (PO) if there exists no other outcome $o'$ such that $o'$ is at least as preferred as $o$ or incomparable with $o$ for all agents, and strictly preferred to $o$ for at least one agent.*

Pareto-optimality (PO) implies weak Pareto-optimality (WPO). That is, when an alternative is PO, it is also WPO. However, the reverse does not hold: a WPO alternative is not necessarily PO.

**Definition 2** (Rational agent). *A rational agent $i$ will accept an agreement $o'$ instead of $o$ only if $o'$ is at least as preferred as $o$ ($o' \succeq_i o$) or incomparable with $o$ ($o' \bowtie_i o$).*

That means, an rational agent $i$ will accept an agreement $o'$ instead of $o$ only if $o$ is not strictly preferred to $o'$.

### 3.1 The framework

**Negotiation tree**

For a negotiation problem over $m$ attributes $\mathbf{V} = \{X_1, \ldots, X_m\}$, we conceptualize the assignment of the

---
of non-dominated outcomes among the feasible outcome space with respect to an agent's preferences.

attribute values as a tree, known as the *negotiation tree*. Let $k$ be the maximum size of the attribute domain: $\forall X \in \mathbf{V},\ |D(X)| \leq k$, the negotiation tree is then a $k$-ary tree. The depth of the negotiation tree is $m$ with the root being at depth 0. We assume that the set of attributes are ordered in some way $\mathcal{O} = X_{\sigma_1} > \cdots > X_{\sigma_m}$, e.g., a random order chosen by a non-bias natural device. The root node represents an empty assignment; each *path* to a node at depth $\ell$ specifies a unique value assignment $assg \in D(X_{\sigma_1}) \times \cdots \times D(X_{\sigma_\ell})$ to the set of attributes $\{X_{\sigma_1}, \ldots, X_{\sigma_\ell}\}$ in that order. Each node at depth $m$ represents one possible alternative (outcome) and the path to reach that node from the root specifies the complete assignment to the set of domain attributes according to that alternative. Such a negotiation tree is shown in Figure 2 for a bilateral negotiation scenario over a set of three attributes $\mathbf{V} = \{A, B, C\}$, where the agents preferences are presented by the CP-nets depicted in Figure 1.

A negotiation tree is created iteratively by the negotiating agents in a distributed manner and under incomplete information setting. In each iteration, the only information that each agent obtains is the nodes in the negotiation tree which are currently available for him to make a proposal on. There is no prior information about the preferences of the opponents. Moreover, the proposals made by an agent during negotiation are *invisible* for its opponents. The procedure starts with a root with an empty assignment. The negotiation tree is then created in a top-down process, where in each iteration of negotiation, each agent can only choose one of the existing *leaf* nodes in the negotiation tree to make a proposal on. Note that the negotiating agents may make proposals on different nodes in an iteration of negotiation. Note also that each agent is not allowed to make proposals on the same node more than once during the entire negotiation process. Once a leaf node $\eta$ at depth $\ell$ ($\ell < m$) is agreed by all the participating agents (i.e., every agent has at some point made a proposal on that node during negotiation), the subtree of $\eta$ will be expanded with every possible value assigns to the next attribute $X_{\sigma_{\ell+1}}$; and these children nodes will be explored by the agents and be available for them to make proposals on in the next iteration. We formally describe the following definitions of *open nodes* and *agreement nodes* in a negotiation tree.

**Definition 3** (Open node). *A node $\eta$ in the negotiation tree is marked as open if and only if it is a* leaf *node and it is agreed by all the negotiation agents, i.e., every agent has ever made a proposal on this node during negotiation (not necessarily in the current iteration).*

Note that once a node $\eta$ at depth $\ell$ ($\ell < m$) is marked as open in the current iteration, it will be expanded with every possible value assigns to the next attribute $X_{\sigma_{\ell+1}}$. Thus, in the next iteration, $\eta$ is not an open node any more, because it will no longer be a leaf node.

**Definition 4** (Agreement node). *A node $\eta$ in the negotiation tree is an agreement node if and only if it is an* open *node at depth $m$.*

An initial agreement is reached if there is at least one agreement node in the negotiation tree. The path to reach an agreement node from the root specifies the complete assignment to the set of domain attributes that the agents have agreed on during negotiation.

### Best possible agreement (BPA)

At each node $\eta$ of the negotiation tree, each agent $i$ has a best possible agreement on that node, denoted by $BPA_i(\eta)$, which is the optimistic outcome that agent $i$ can obtain with the values assigned to the attributes along the path from the root to $\eta$ being fixed. Let $assg = \texttt{PATH}(\eta)$ be the value assignment specified by the path from the root to $\eta$, then $BPA_i(\eta)$ is the best outcome among the completions of $assg$ ($Comp(assg)$) for agent $i$. Moreover, the best possible agreement (BPA) of the root node for an agent corresponds to the optimal (best) alternative of that agent in the entire outcome space, i.e. each attribute is assigned a most preferred value according to that agent's preference.

In the context of acyclic CP-nets, computing the best possible agreement of a node for an agent is similar to the individual outcome optimization with constraints (i.e., the values assigned to the attributes along the path from the root node to the current node being fixed) [1]. We simply need to sweep through the network from ancestors to descendants, assigning the most preferred value to each remaining attribute $X$ (i.e., the attribute that has not been assigned a value along the path) respecting to the parent context. For instance, consider the agents' CP-nets in Figure 1 and assume a path assignment for a node $\eta$ is $a$. According to agent 1's CP-net in Figure 1(a), we consider an order over attributes from ancestors to descendants: $\mathcal{O} = C > A > B$. We first assign $c$ to $C$, because $c \succ \bar{c}$. The next variable to be considered is $B$, because the value of $A$ has already been specified by the path to $\eta$. Then we assign $b$ to $B$, because $b \succ \bar{b}$ given $A = a$ and $C = c$. Consequently, $BPA_1(\eta) = abc$. Similarly, for agent 2, $BPA_2(\eta) = a\bar{b}\bar{c}$.

### 3.2 The process of negotiation

We now present an example of a bilateral negotiation scenario using POANCD in combinatorial domains. POANCD is defined in two phases. The first phase of

`POANCD` consists of distributed formation of a negotiation tree by the participating agents. After the first phase, the agents will be left with a few initial agreements. In the second phase, the agents will act cooperatively to achieve Pareto-optimal agreement by exploring possible mutually beneficial alternatives.

**First phase of `POANCD`:**

**Step 1:** A random device chooses an order over the domain attributes, e.g., $\mathcal{O} = X_{\sigma_1} > \cdots > X_{\sigma_m}$, such that the negotiation tree is created following that order. Initially, a root node and all its possible $|D(X_{\sigma_1})|$ children nodes (each branch assigns a distinct value to the attribute $X_{\sigma_1}$) are created in the negotiation tree.

**Step 2:** Each negotiating agent makes a proposal on an existing *leaf* node in the negotiation tree.

After each agent makes a proposal, let $\mathbf{Q}$ denotes the set of nodes marked as open in the current iteration. Note that there would be at most two nodes marked as open in each iteration ($|\mathbf{Q}| \leq 2$), because we are considering a bilateral negotiation and in each iteration, each agent can only make a proposal on one node [4].

- If there exist at least one agreement nodes in the negotiation tree, collect the set of *open* nodes $\mathbf{Q}$ and go to Step 3.
- Otherwise, for each $\eta \in \mathbf{Q}$, let $\ell$ denotes the depth of $\eta$, and thus the next attribute to be considered in the subtree of $\eta$ is $X_{\sigma_{\ell+1}}$. Expand $\eta$ with all possible $|D(X_{\sigma_{\ell+1}})|$ children nodes in the negotiation tree; and go back to Step 2.

REMARK. The negotiation process takes place under incomplete information setting and in a distributive environment. In each iteration of negotiation, each agent $i$ only knows a set of nodes (options), denoted by $\Omega_i$, on which he can make proposals, i.e., the set of leaf nodes in the negotiation tree that he has not made a proposal on during the previously rounds. For each node $\eta$ in $\Omega_i$, an agent $i$ will have a best possible agreement (BPA) of $\eta$ ($BPA_i(\eta)$), which indicates the optimistic outcome that agent $i$ can obtain following the subtree of $\eta$. An agent can always choose to go backward as long as the BPA of the current node is less preferred than that of another node. Thus, a rational agent will always try to get the most preferred alternative among the possible options. Consequently, we consider the following strategy in negotiation: chooses a node $\eta \in \Omega_i$ whose BPA is the best among the BPAs of the nodes in $\Omega_i$. That means, there does *not* exist another node $\eta'$ in $\Omega_i$ ($\eta' \neq \eta$), such that $BPA_i(\eta') \succ_i BPA_i(\eta)$.

As an implementation, each agent can maintain a priority queue of feasible leaf nodes to make proposals on, known as the $fringe$. The nodes in an agent's $fringe$ is sorted according to the preference ordering over the BPAs of these nodes for that agent. The more preferred the $BPA_i(\eta)$ is, the higher priority the node $\eta$ is in the $fringe$ of agent $i$. In each iteration of negotiation, for each agent $i$, the first node is removed from the $fringe$ and agent $i$ will make a proposal in the negotiation tree on that node. If there are new nodes created in the negotiation tree (i.e., the children nodes of the open nodes created in the current iteration), each agent will add the new nodes into its $fringe$ according to its own preference ordering on the BPAs. Figure 3 shows an example of two participating agents' $fringe$s in a bilateral negotiation corresponding to the negotiation scenario depicted in Figure 2. Note that with acyclic CP-nets, the BPA is always unique at each node of the negotiation tree. With other types of preference when there may exist more than one best possible agreements, the BPA can be defined as a set of most preferred outcomes, which will be all added into the $fringe$. The $fringe$ will be sorted and in the next iteration the agent will choose the first one to make an offer on. Finally notice that even if the BPA of a node is unique, the $fringe$ can also contain incomparable outcomes, i.e., the BPA of different nodes in the tree might be incomparable.

**Step 3:** We refer to the set of agreement nodes as $\mathbf{A}$ and the set of agreements (complete assignments) corresponding to the agreement nodes in $\mathbf{A}$ as $\mathbf{I}$: $\mathbf{I} = \{o^* \mid o^* = \texttt{PATH}(\eta^*) \text{ and } \eta^* \in \mathbf{A}\}$, where $\texttt{PATH}(\eta^*)$ denotes the value assignment to the set of domain attributes specified by the path from the root to $\eta^*$. Note that an agreement node is also an open node, thus $\mathbf{A} \subseteq \mathbf{Q}$.

- If $\mathbf{Q} = \mathbf{A}$, the first phase ends and proceed to the second phase.
- Otherwise, the set $\mathbf{Q}$ must contain one agreement node at depth $m$, denoted by $\eta^*$; and one open node at depth $\ell$ ($\ell < m$), denoted by $\eta'$. That means, in the last iteration, an agent $i$ makes a proposal on $\eta^*$, which has been agreed by the other agent $j$ ($i \neq j$) during a previous iteration of negotiation; and the other agent $j$ makes a proposal on $\eta'$, which has been agreed by agent $i$ in a previous iteration of negotiation. Let $o^*$ be the complete assignment (alternative) specified by the path from the root to the agreement node $\eta^*$. Even though $\eta'$ is not an agreement node, there does exist a potential agreement under the subtree of $\eta'$, because both agents have agreed on $\eta'$. Moreover, this potential agreement can not be strictly preferred to $o^*$ for agent $j$ (otherwise, he would have made a proposal on $\eta'$ before he makes a proposal $\eta^*$), but may be more preferred than $o^*$ to agent $i$. However, since there already exists an agreement $o^*$, agent $j$ will not make further concession in the subtree of $\eta'$. Consequently, in order to be fair, we ask agent $j$, who proposes $\eta'$ in the last iteration, to give out the BPA of $\eta'$ ($BPA_j(\eta')$). Then agent $i$ can either choose to include $BPA_j(\eta')$ in the initial agreement set: $\mathbf{I} = \mathbf{I} \cup \{BPA_j(\eta')\}$; or stick with the current set $\mathbf{I}$. Note that agent $i$ will choose to include $BPA_j(\eta')$ only if $o^* \not\succ_i BPA_j(\eta')$. The first phase ends and proceed to the second phase with a set of initial agreements $\mathbf{I}$.

---
[4] In a multilateral negotiation, there may be more than two nodes marked as open in an iteration of negotiation.

**Second phase of `POANCD` (Enhancement):**

This phase is also called the *enhancement phase*, in which the participating agents will act cooperatively to explore possible mutually beneficial agreements and decide on the final agreement. We first introduce the following notations:

1. let $\Omega_i$ denote the set of leaf nodes in the negotiation tree that agent $i$ has not yet made a proposal on during the first phase of negotiation, i.e., the remaining nodes in agent $i$'s $fringe$;

2. For each initial agreement $o^* \in \mathbf{I}$ and each negotiating agent $i$, let:
   - $\Gamma_i(o^*)$ denote the set of nodes in $\Omega_i$ ($\Gamma_i(o^*) \in \Omega_i$) whose BPAs are indifferent or incomparable with $o^*$ for agent $i$: $\Gamma_i(o^*) = \{\eta \in \Omega_i \mid BPA_i(\eta) \sim_i o^*$ or $BPA_i(\eta) \bowtie_i o^*\}$; and
   - $\Theta_i(o^*)$ denote the corresponding set of agent $i$'s BPAs of the nodes in $\Gamma_i(o^*)$: $\Theta_i(o^*) = \{o \mid o = BPA_i(\eta)$ and $\eta \in \Gamma_i(o^*)\}$.

Notice that for any node $\eta$ in $\Omega_i$, $BPA_i(\eta)$ can not be strictly preferred to $o^*$ for agent $i$. Otherwise, agent $i$ would have made a proposal at $\eta$ before he makes a proposal on the corresponding agreement node of $o^*$.

In this phase, for each initial agreement $o^*$, let us define a set $\Phi(o^*)$, which is the set of Pareto-optimal alternatives that can possibly replace $o^*$ in $\mathbf{I}$. Originally, $\Phi(o^*) = \emptyset$. Each agent $i$ gives out $\Theta_i(o^*)$, then the other agent (agent $j$) can either choose one of the alternatives in $\Theta_i(o^*)$ to put in $\Phi(o^*)$ or stick with $o^*$. Note that agent $j$ will choose an alternative $o'$ from $\Theta_i(o^*)$ to put in $\Phi(o^*)$ only if $o^* \not\succ_j o'$; and the alternative $o'$ that agent $j$ chooses would be the best alternative among $\Theta_i(o^*)$ for agent $j$. If $\Phi(o^*) \neq \emptyset$, then $o^*$ in $\mathbf{I}$ will be replaced by the outcomes in $\Phi(o^*)$.

Finally, after the second phase, the set $\mathbf{I}$ contains the set of final agreements. If only one element remains in $\mathbf{I}$, it will be selected as the final agreement. Otherwise, any one of them will be chosen randomly as the final agreement.

### 3.3 Formal properties of `POANCD`

In this section, we discuss the formal properties of the proposed `POANCD` negotiation protocol. The objective of an efficient protocol is to lead rational agents to efficient agreements [9]. We present the formal proof of Pareto-optimality of our proposed protocol as follows.

**Theorem 1.** *The agreements reached by `POANCD` is Pareto-optimal.*

In order to proof this theorem, we first need to proof the following proposition.

**Proposition 1.** *The initial agreements reached in the first phase of `POANCD` is weakly Pareto-optimal.*

*Proof.* Assume first that an initial agreement $o^*$ on node $\eta^*$ that the agents reach in the first phase of `POANCD` is not weakly Pareto-optimal, then there exists another alternative $o'$, such that for any agent $i$: $o' \succ_i o^*$. We assume $\eta'$ is a *leaf* node whose path assignment is coincided with $o'$. For any agent $i$, $BPA_i(\eta') \succeq_i o'$ and thus $BPA_i(\eta') \succ_i o^*$. Then both agents will make the proposals on $\eta'$ or the nodes in the subtree of $\eta'$ before they make proposals on $\eta^*$; and $\eta^*$ will not be an agreement node, contradicting the face that $o^*$ is an initial agreements from the first phase of negotiation. □

*Proof of Theorem 1.* From Proposition 1 we know that the initial agreements the agents reach in the first phase of negotiation is weakly Pareto-optimal. In the second phase, the agents are acting cooperatively to reach Pareto-optimal agreements by replacing each inefficient initial agreement $o^*$ with a set of Pareto-optimal alternatives $\Phi(o^*)$, such that every alternative $o' \in \Phi(o^*)$ is indifferent or incomparable with $o^*$ for one agent $i$, and is more preferred than $o^*$ for the other agent $j$. Moreover, after an agent $i$ gives out the set $\Theta_i(o^*)$, if the other agent $j$ would like to replace $o^*$ in $\mathbf{I}$, the alternative he chooses from $\Theta_i(o^*)$ will be

the best one among $\Theta_i(o^*)$. Consequently, the final agreement reached by POANCD is Pareto-optimal. □

### 3.4 An illustration

Now, we demonstrate the execution of POANCD with an example. Assume two agents are negotiating over a set of three binary-valued attributes $\mathbf{V} = \{A, B, C\}$. The negotiating agents' preferences are depicted in Figure 1. As all attributes are binary, the negotiation tree is a binary tree. In the first phase, firstly, an ordering over the attributes is randomly generated, e.g., we consider the ordering $\mathcal{O} = A > B > C$ following which the negotiation tree will be created. Each node $\eta$ in the negotiation tree associates with a proposal table, in which the first row displays the proposals that the agents make on that node: the left (resp. right) column depicts the proposal that agent 1 (resp. agent 2) makes; each proposal is marked with a number that depicts the number of the current iteration, i.e., a proposal marked by "(p)" (resp. "<p>"), is the proposal that agent 1 (resp. agent 2) makes in the $p^{th}$ iteration. For explanation purpose, we also attach the best possible agreements (BPA) of both agents at each node in the second row of the table: the left (resp. right) column depicts the BPA of that node of agent 1 (resp. agent 2). However, it is important to note that the information including the proposal that an agent makes and the BPA of a node of that agent is its private information and invisible for its opponent. Figure 2 shows the formation of the negotiation tree in the first phase of POANCD and Figure 3 provides an illustration of the ongoing changes occurs in each agent's $fringe$ [5]. For the purpose of explanation, we also provide both participating agents' preference orderings over the outcome space in Figure 4, which are induced from the corresponding agents' CP-nets in Figure 1. However, note that the agents do not need to reason about the preference relations over the entire outcome space during negotiation. They only need to answer a few dominance queries when adding new nodes into their $fringe$s.

Initially, a root node is created in the negotiation tree. Since the first attribute to be considered is $A$ and $D|A| = \{a, \bar{a}\}$, two children nodes $\eta_1$, $\eta_2$ from the branch $a$, $\bar{a}$ are created in the negotiation tree. Each agent will create a $fringe$ and add $\eta_1$ and $\eta_2$ into its $fringe$ according to their preference orderings on the BPAs of $\eta_1$ and $\eta_2$. For instance, $BPA_1(\eta_1) = abc$, $BPA_1(\eta_2) = \bar{a}\bar{b}c$, because $abc \succ_1 \bar{a}\bar{b}c$ (see the preference ordering of agent 1 in Figure 4(a)), the order in agent 1's $fringe$ is $\eta_1\eta_2$. Similarly, the order in agent 2's $fringe$ is $\eta_1\eta_2$.

---
[5]The nodes depicts in red colour are the new nodes created in that iteration.

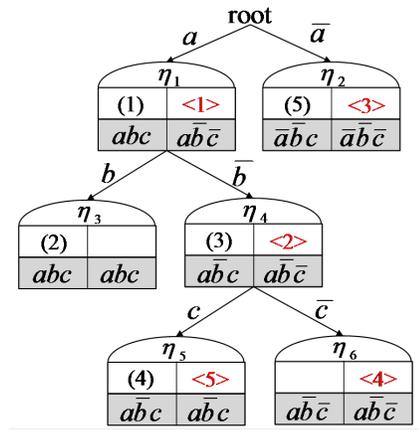

Figure 2: Negotiation tree

In the $1^{st}$ iteration, each agent will make a proposal on one of the leaf nodes in the negotiation tree. For instance, the first node $\eta_1$ in agent 1's $fringe$ is pop out and agent 1 makes a proposal (1) at $\eta_1$ in the negotiation tree. Similarly, agent 2 also makes a proposal <1> on node $\eta_1$. $\eta_1$ is marked as open in this iteration and $\mathbf{Q} = \{\eta_1\}$. Since the next attribute to be considered in the subtree of $\eta_1$ is $B$ and $D(B) = \{b, \bar{b}\}$. Two children nodes $\eta_3$, $\eta_4$ of $\eta_1$ from the branches $b$, $\bar{b}$ respectively are created in the negotiation tree. Each agent adds these two nodes into its own $fringe$ according to its preference. For instance, $BPA_1(\eta_2) = \bar{a}\bar{b}c$, $BPA_1(\eta_3) = abc$, $BPA_1(\eta_4) = a\bar{b}c$, and because $abc \succ_1 a\bar{b}c \succ_1 \bar{a}\bar{b}c$, the $fringe$ of agent 1 is $\eta_3\eta_4\eta_2$ (Figure 3(a)); similarly the $fringe$ of agent 2 is $\eta_4\eta_2\eta_3$ (Figure 3(b)).

In the $2^{nd}$ iteration, agent 1 makes a proposal (2) on node $\eta_3$ and agent 2 makes a proposal <2> on $\eta_4$. In this iteration, there is no open node ($\mathbf{Q} = \emptyset$); agent 1's $fringe$ is $\eta_4\eta_2$ and agent 2's $fringe$ is $\eta_2\eta_3$.

In the $3^{rd}$ iteration, agent 1 continues to make a proposal (3) on $\eta_4$ (i.e., the first node in its $fringe$) and agent 2 makes a proposal <3> on node $\eta_2$. There is one node marked as open in the current negotiation tree: $\mathbf{Q} = \{\eta_4\}$, thus we created two children nodes $\eta_5$, $\eta_6$ of $\eta_4$ from the branches assign $c$ and $\bar{c}$ to the next attribute $C$ respectively. Both agents add these two nodes into their $fringe$s for consideration according to their preferences over the BPAs of the nodes: agent 1's $fringe$ becomes $\eta_5\eta_2\eta_6$ and agent 2's $fringe$ becomes $\eta_6\eta_5\eta_3$.

In the $4^{th}$ iteration, agent 1 makes a proposal (4) on $\eta_5$ and agent 2 makes a proposal <4> on node $\eta_6$. No node is marked as open in this iteration of negotiation; agent 1's $fringe$ is $\eta_2\eta_6$ and agent 2's $fringe$ is $\eta_5\eta_3$. Finally, in the last iteration (the $5^{th}$ iteration), agent 1 makes a proposal (5) on node $\eta_2$ and agent 2 makes a proposal <5> on node $\eta_5$. As such, in this iteration, there are two nodes marked as open: $\mathbf{Q} = \{\eta_2, \eta_5\}$.

| Initial | $\eta_1$ | $\eta_2$ | |
|---|---|---|---|
| 1st iteration | $\eta_3$ | $\eta_4$ | $\eta_2$ |
| 2nd iteration | $\eta_4$ | $\eta_2$ | |
| 3rd iteration | $\eta_5$ | $\eta_2$ | $\eta_6$ |
| 4th iteration | $\eta_2$ | $\eta_6$ | |
| 5th iteration | $\eta_6$ | | |

(a) Agent 1

| Initial | $\eta_1$ | $\eta_2$ | |
|---|---|---|---|
| 1st iteration | $\eta_4$ | $\eta_2$ | $\eta_3$ |
| 2nd iteration | $\eta_2$ | $\eta_3$ | |
| 3rd iteration | $\eta_6$ | $\eta_5$ | $\eta_3$ |
| 4th iteration | $\eta_5$ | $\eta_3$ | |
| 5th iteration | $\eta_3$ | | |

(b) Agent 2

Figure 3: The fringes of the agents in the negotiation

$$abc \succ a\bar{b}c \succ \bar{a}\bar{b}c \quad \begin{array}{l} \succ \bar{a}bc \\ \succ \bar{a}b\bar{c} \\ \succ \bar{a}\bar{b}\bar{c} \\ \succ a\bar{b}\bar{c} \end{array} \succ ab\bar{c}$$

(a) Agent 1

$$a\bar{b}\bar{c} \succ \bar{a}\bar{b}\bar{c} \succ \bar{a}\bar{b}c \succ a\bar{b}c \succ abc \succ \bar{a}bc \succ \bar{a}b\bar{c} \succ ab\bar{c}$$

(b) Agent 2

Figure 4: The preference orderings of two agents

Moreover, since $\eta_5$ is an agreement node (i.e., it is at depth 3), Step 2 ends and we proceed to Step 3 of the first phase.

As $\mathbf{A} = \{\eta_5\}$ and $\mathbf{A} \neq \mathbf{Q}$, $\mathbf{Q}$ contains one agreement node $\eta_5$ (the path from the root to $\eta_5$ specifies a complete assignment $a\bar{b}c$); and one open node $\eta_2$ (the path from the root to $\eta_2$ specifies a partial assignment $\bar{a}$). Since agent 1 is the proposer of node $\eta_2$ in the last iteration, it gives out its BPA of $\eta_2$: $BPA_1(\eta_2) = \bar{a}\bar{b}c$. For agent 2, as $\bar{a}\bar{b}c \succ_2 a\bar{b}c$ (see the preference ordering of agent 2 in Figure 4(b)), agent 2 will include $\bar{a}\bar{b}c$ in the set of initial agreements. Consequently, the first phase ends and we proceed to the second phase of POANCD with the set of initial agreements $\mathbf{I} = \{a\bar{b}c, \bar{a}\bar{b}c\}$.

Originally, $\mathbf{F} = \{a\bar{b}c, \bar{a}\bar{b}c\}$. For the initial agreement $a\bar{b}c$, originally $\Phi(a\bar{b}c) = \emptyset$. For agent 1, there is only one leaf node $\eta_6$ that he has not yet made a proposal on ($\Omega_1 = \{\eta_6\}$) and $BPA_1(\eta_6) = a\bar{b}\bar{c}$. Since $a\bar{b}c \succ_1 a\bar{b}\bar{c}$, there is no node in $\Omega_1$ whose BPA is indifference or incomparable with $a\bar{b}c$ for agent 1, $\Gamma_1(a\bar{b}c) = \emptyset$ and thus $\Theta_1(a\bar{b}c) = \emptyset$. Similarly, for agent 2 whose CP-net induced a strict total preference ordering over the outcome space (see Figure 4(b)), the BPAs of the leaf nodes that he has not yet made proposals on are less preferred than the current agreement $a\bar{b}c$. Hence, $\Gamma_2(a\bar{b}c) = \emptyset$ and $\Theta_2(a\bar{b}c) = \emptyset$. Consequently, $\Phi(a\bar{b}c) = \emptyset$ and $a\bar{b}c$ will not be replaced. Similarly, for another initial agreement $\bar{a}\bar{b}c$, $\Phi(\bar{a}\bar{b}c) = \emptyset$ and $\bar{a}\bar{b}c$ will not be replaced. Consequently, both initial agreements $a\bar{b}c$ and $\bar{a}\bar{b}c$ are Pareto-optimal and we obtain the set of final agreements $\mathbf{F} = \{a\bar{b}c, \bar{a}\bar{b}c\}$. As $\mathbf{F}$ contains more than one element, we randomly select one of them as the final agreement and the negotiation process ends.

## 4 Experiment

We now describe the results of experiments that show the feasibility and computational efficiency of our proposed POANCD protocol to bilateral negotiation in combinatorial domains with respect to *(i)* the number of attributes $s_{attr}$ and domain size $s_{ds}$ that can be efficiently handled in practice; *(ii)* the corresponding outcome space $s_{os}$ and the average number of different outcomes (alternatives) $s_{out}$ that each agent needs to consider during the entire negotiation process; *(iii)* the average number of dominance queries (outcome comparisons) $s_{dq}$ that each agent needs to answer during the entire negotiation process; *(iv)* the average number of iterations $s_{iter}$ that the first phase of the negotiation process involves; and *(v)* the average execution time $s_{time}$ of the entire negotiation process.

In these experiments, we use CP-nets for representing the agents' preferences in negotiation. We consider a simple CP-net structure, i.e., directed-path singly connected CP-nets; and we restrict the maximum in-degree of each node in the generated CP-nets to 5. We generate random preference networks by varying the number of attributes, the structure of the network and the preferences for the attributes. In the negotiation process, we use the individual outcome optimization technique in CP-nets introduced in [1] and the heuristic approach to answer dominance query in CP-nets introduced by Li *et al.* [6].

We first conduct the experiments for binary-valued CP-nets, in which the number of attributes $s_{attr}$ is varying from 2 to 100 and for each number of attributes we run 1000 rounds of experiments by randomly generating two negotiating agents' preferences. Table 1 shows the experimental result in binary CP-nets. The average number of outcomes $s_{out}$ that each agent needs to consider, the average number of dominance queries $s_{dq}$, the average number of iterations $s_{iter}$ and the average execution time $s_{time}$ are increasing as the number of attributes increases. However, we can observe that the proposed negotiation protocol POANCD can efficiently handle large number of attributes in negotiation with binary CP-nets. Compared to the huge outcome space, by using the proposed protocol POANCD, the average number of alternatives $s_{out}$ that each agent needs to consider is significantly reduced (comparing the second and third column of Table 1). When the number of attributes is large (e.g., 100), on average, the number of dominance queries $s_{dq}$ that each agent needs to answer during the entire negotiation process is less than 420 and the first phase of negotiation finishes in less than 170 rounds. Moreover, according to the experiment data, when the number of attributes is 100, on average, the entire negotiation process ends in about 30 seconds.

Table 1: Negotiations with binary CP-nets

| $s_{attr}$ | $s_{os}$ | $s_{out}$ | $s_{dq}$ | $s_{iter}$ | $s_{time}$ (sec) |
|---|---|---|---|---|---|
| 2 | 4 | 2.80 | 5.36 | 3.38 | 0.10 |
| 10 | 1024 | 15.84 | 45.05 | 17.70 | 0.30 |
| 20 | 1048576 | 31.19 | 86.03 | 34.31 | 1.06 |
| 50 | $2^{50}$ | 77.67 | 207.42 | 84.28 | 6.60 |
| 100 | $2^{100}$ | 155.14 | 410.62 | 168.09 | 30.03 |

Table 2: Negotiations with multi-valued CP-nets

| $s_{attr}$ | $s_{os}$ | $s_{out}$ | $s_{dq}$ | $s_{iter}$ | $s_{time}$ (sec) |
|---|---|---|---|---|---|
| 2 | 25 | 2.81 | 17.24 | 4.17 | 0.05 |
| 5 | 3125 | 8.56 | 70.18 | 12.21 | 0.78 |
| 10 | 9765625 | 16.06 | 125.69 | 22.86 | 9.81 |
| 15 | $2^{15} \sim 5^{15}$ | 22.71 | 169.99 | 32.75 | 151.20 |
| 25 | $2^{25} \sim 5^{25}$ | 37.75 | 266.62 | 53.06 | 301.05 |

We extend the experiment for multi-value CP-nets (see Table 2). For multi-valued CP-nets, we restrict the maximum domain size to 5. We vary the number of attributes $s_{attr}$ from 2 to 25 and for each number of attributes we run 500 rounds of experiments. Similar to the scenario with binary-valued CP-nets, $s_{out}$, $s_{dq}$, $s_{iter}$ and $s_{time}$ are increasing as the number of attributes increases. However, although with 25 attributes (in which case the maximum domain size is $5^{25}$), on average, each agent only needs to consider less than 40 alternatives and answer less than 300 dominance queries during the whole negotiation process; the first phase of POANCD finishes in about 50 iterations. Moreover, according to the experimental data, on average the entire negotiation process ends in about 300 seconds with 25 attributes. Note that in multi-value CP-nets, answering dominance queries is much more complex than that in binary-valued CP-nets.

From the experiment, we can conclude that our proposed POANCD protocol is computationally efficient. It allows preferences structures that are quite large and complex to be executed in reasonable time.

## 5 Conclusion and Future Work

In this paper, we propose an efficient distributed negotiation protocol, POANCD, for negotiation in combinatorial domains when the agents do not know the preferences of each other and they do not want to reveal their preferences for the possible alternatives during the process of negotiation. We have theoretically shown that POANCD leads rational agents to Pareto-optimal agreements. We have also experimentally shown the significant reduction of search efforts and the number of dominance queries each participating agent needs to answer by using the proposed protocol.

A major advantage of POANCD is its extensibility to multilateral negotiation. We have presented POANCD for bilateral negotiation, but extension to multilateral negotiation can be done with minor modifications.

However, the negotiation scenarios with cyclic or inconsistent preferences need to be further explored, because there may be more than one best possible agreement (BPA) of a node in a negotiation tree. Moreover, the fairness issue of the negotiated outcome using the proposed protocol is also an important aspect of future research.

## Acknowledgements

This work is partially supported by the ARC Discovery Grant DP110103671.